\documentclass{article}

\PassOptionsToPackage{numbers, compress}{natbib}
\usepackage[final]{nips_2017}

\usepackage[utf8]{inputenc} 
\usepackage[T1]{fontenc}    
\usepackage{hyperref}       
\usepackage{url}            
\usepackage{booktabs}       
\usepackage{amsfonts}       
\usepackage{nicefrac}       
\usepackage{microtype}      
\usepackage{graphicx}
\usepackage{amsmath}

\title{Predicting Adolescent Suicide Attempts with Neural Networks}

\author{
  Harish S.~Bhat\\
  Applied Mathematics Unit\\
  University of California, Merced\\
  Merced, CA 95343\\
  \texttt{hbhat@ucmerced.edu} \\
  \And
  Sidra J. Goldman-Mellor\\
  Public Health\\
  University of California, Merced\\
  Merced, CA 95343\\
  \texttt{sgoldman-mellor@ucmerced.edu}\\
}

\begin{document}

\maketitle

\begin{abstract}
Though suicide is a major public health problem in the US, machine learning methods are not commonly used to predict an individual's risk of attempting/committing suicide. In the present work, starting with an anonymized collection of electronic health records for 522,056 unique, California-resident adolescents, we develop neural network models to predict suicide attempts. We frame the problem as a binary classification problem in which we use a patient's data from 2006-2009 to predict either the presence (1) or absence (0) of a suicide attempt in 2010. After addressing issues such as severely imbalanced classes and the variable length of a patient's history, we build neural networks with depths varying from two to eight hidden layers. For test set observations where we have at least five ED/hospital visits' worth of data on a patient, our depth-4 model achieves a sensitivity of 0.703, specificity of 0.980, and AUC of 0.958. 
\end{abstract}

\section{Introduction}
Recently, both suicide fatalities and attempts have increased among US adolescents.  Between 2009 and 2014, suicide deaths among youths aged 10-19 increased 17\%, and nonfatal suicide attempts that resulted in an emergency department (ED) visit soared 65\% \citep{CDC2013, pittsenbarger2014}.  At the same time, the predictive capability of clinicians and standard statistical models is only slightly better than chance \citep{franklin2017}. Given the ever increasing volume of electronic health records (EHRs), it is natural to ask whether machine learning methods might be capable of improving the accuracy of suicide attempt prediction.  If so, such methods might one day be incorporated into clinical tools that flag high-risk patients, potentially improving their lives.

Modern deep learning methods have not made their way into the mainstream suicide prediction literature.  Recent work features methods such as random forests \citep{walsh2017} and $k$-nearest neighbors \citep{tran2013}.  Other studies involve special data sources, such as textual notes in EHRs \citep{haerian2012}, survey data \citep{delgadogomez2011, delgadogomez2012, ruiz2012, zaher2016}, or professional risk assessments \citep{tran2013}. This enriched data enables the use of methods from, e.g., natural language processing and ordinal regression.  Outside of suicide research, deep learning has enjoyed recent success in the medical informatics literature \citep{che2016, ravi2017}, further motivating the present work.

We have found only two prior papers in which neural network methods of any kind have been used in suicide research.  In \citet{iliou2016}, a combination of \emph{ad hoc} feature selection methods and machine learning methods (including multilayer perceptrons) are applied to data from $91$ patients. In \citet{nguyen2016}, deep neural nets with dropout are applied to data from $7399$ patients.  However, both prior studies \citep{iliou2016, nguyen2016} focus exclusively on patients with a prior history of depression and/or other mental health conditions.  In the latter study, the set of predictors includes a suicide risk assessment score for each patient \citep{tran2013, nguyen2016}.  The problem with focusing on such data sets is that more than 50\% of suicide decedents have no record of psychiatric disorders, psychiatric care-seeking, or suicidality \citep{Husky2012, Chock2015, Ahmedani2014}.  The present work is the first to use electronic health records from a broad subset of the population, unrestricted by prior conditions/diagnoses, to build neural network models to predict suicide attempts.

\section{Data}
\label{sect:data}
This study was approved by UC Merced's Institutional Review Board. We used nonpublic versions of California emergency department encounter and hospital admissions data from 2006 through 2010. The California Office of Statewide Health Planning and Development (OSHPD) provided anonymized individual-level patient encounter data from all California-licensed hospital facilities, including general acute care, acute psychiatric, chemical dependency recovery, and psychiatric health facilities, but excluding federal hospitals. ED and inpatient data were screened by the OSHPD's automated data entry and reporting software program (MIRCal); data fields with error rates of 0.1\% or higher were returned to the hospitals for correction \citep{oshpd2017a, oshpd2017b}. Patients with missing age were excluded.

The study dataset consists of ED and inpatient records for all adolescent patients aged 10 to 19 years who had a valid unique identifier (encrypted social security number or SSN) and a California residential zip code in 2010 (64.0\% of all records for this age group). Unique identifiers were used to link multiple ED visits/episodes per patient over time, including encounters prior (2006-2009) to the adolescent's first recorded 2010 visit. A total of 522,056 unique, CA-resident adolescents are available for these analyses. Of these, 5,490 adolescent patients presented in 2010 to an ED with a suicide attempt code, i.e., a primary International Classification of Diseases, Ninth Revision, Clinical Modification (ICD-9-CM \citep{medicode}) External Cause of Injury code (E-code) of E950.0-959.x \citep{crosby2011}.

Note that prior neural network suicide prediction models were trained using data from, respectively, $91$ and $7399$ patients, nearly $70$ times less than the present work \citep{iliou2016, nguyen2016}.  Significantly, these and many other machine learning studies restrict their attention to data where all patients have a history of mental illness.  Our long-term goal is to build models that can compute individualized risk of suicide attempt/fatality for all ED patients; naturally, this includes those with and without a history of any particular illness.  Hence our data set is much larger in size and scope than in prior studies.

Note that we do not have linked death records.  While we know that the data includes both fatal and non-fatal suicide attempts, with rare exceptions we cannot tell from the present data set whether a particular attempt was fatal.  

In this work, we treat each ED/hospital visit in 2006-2009 by each patient as a separate observation.  There are 772,923 such visits in the full data set.  To each visit, we assign an outcome variable $y^f$ of either $1$ (presented in 2010 with a suicide attempt code, as defined above) or $0$ (absence of suicide attempt code in 2010).  For each visit, the raw predictors include the patient's sex, age, race, insurance category, zip code, county of residence, etc.  For a full list, please see Table \ref{tab:predictors} in the Appendix.  Importantly, our data does not include textual annotations from medical professionals, survey responses, or numerical mental health evaluations.  The vast majority of predictors are categorical; in these cases, we apply one-hot encoding.

The only visit variables that require special treatment are the diagnostic codes associated with the visit.  Each visit is assigned one primary and up to six other Clinical Classifications Software (CCS) diagnostic codes.  The CCS grouping system aggregates >14,000 ICD-9-CM diagnoses into 285 discrete, mutually exclusive, clinically meaningful category codes (e.g., ``anxiety disorders'') that are useful for identifying patients in broad diagnosis groupings.  In our analysis, for each visit, we aggregate all seven one-hot encodings of all CCS diagnostic codes.  The result is a vector $\mathbf{v} \in \mathbb{R}^{285}$ that tracks up to seven diagnoses received in one visit.  

Within the 2006-2009 data corpus, patients may visit one or more times.  To track a patient's history, we include for a visit at time $t$ the cumulative sum of all diagnosis vectors $\mathbf{v}$ (computed as above) recorded at times less than or equal to $t$.  We also record in a new column how many times the patient has been observed in the data set up to and including time $t$.  By virtue of this cumulative encoding, each patient's visit is transformed into a vector of fixed length.  The end result is a full data matrix $X^f$ of size $772923 \times 494$.

We randomly reshuffle the rows of $(X^f,y^f)$ and split the result into a pretraining set $(X, y)$ (80\% of the rows) and a test set $(X^t, y^t)$ (the remaining 20\%).  We record the column means and variances of $X$, discard columns with zero variance (i.e., unobserved diagnostic codes), and normalize all columns of $X$ to have zero mean and unit variance. Note that we are left with $p=382$ columns.

Since only $9804/618338 \approx 1.58\%$ of the $y$ values in the pretraining set are equal to $1$, we have an imbalanced classification problem.  We sample with replacement from the rows of $X$ associated with the minority ($y=1$) class to create a bootstrapped data set with balanced classes.  The resulting data set $(X^b, y^b)$ has $N = 1217068$ rows.

\section{Methods}
As described above, deep neural networks have been underutilized in suicide prediction research.  Hence we focus on neural network classifiers consisting of feedforward networks (equivalently, multilayer perceptrons) with dense, all-to-all connections between layers.  As the mathematical details of such models are standard, we do not include them here; the interested reader should consult the Appendix.  Here we mention the following salient facts.

First, we focus on three networks: NN2, with $d=2$ hidden layers and $50$ units per layer; NN4, with $d=4$ hidden layers and $50$ units per layer; and NN8, with $d=8$ hidden layers and $20$ units in each hidden layer except for the first which has $50$ units.  We chose the NN8 architecture so that the total number of weights is close to the total number of weights in the NN4 model.

Second, we use the scaled exponential linear unit (SELU) $\phi(z) = \lambda (z I_{z > 0}(z) + \alpha (e^z - 1) I_{z \leq 0}(z) )$, with constants $\lambda \approx 1.0507$ and $\alpha \approx 1.6733$ \citep{Klambauer2017}.  This activation function has been engineered to approximately preserve zero-mean, unit-variance normalization across many layers while avoiding vanishing/exploding gradients \citep{Klambauer2017}.  The use of the SELU activation renders batch normalization unnecessary \citep{goodfellow2016}.

Finally, the only regularization technique we use is a variant of early stopping \citep{goodfellow2016}.  From the bootstrapped data set, we make a further split into the training set (80\%) and a validation set (20\%).  We monitor the accuracy, sensitivity, and specificity on both the training and validation sets, and halt optimization when we detect insufficient improvement in our metrics on the validation set.  Importantly, the test set is not used in this procedure.  

\begin{table}[t]
  \centering
  \begin{tabular}{llllllllll}
    \toprule
    Method     & Sens. & Spec. & Prec. & AUC & Method & Sens. & Spec. & Prec. & AUC \\ \midrule
    NN2 & 0.401 & 0.952 & 0.121 & 0.796 & NN2, 662 & 0.710 & \textbf{0.889} & \textbf{0.527} & \textbf{0.896} \\
    NN4 & 0.390 & \textbf{0.956} & \textbf{0.127} & \textbf{0.802} & NN4, 662 & 0.746 & 0.839 & 0.446 & 0.881 \\
    NN8 & \textbf{0.485} & 0.884 & 0.063 & 0.778 & NN8, 662 & \textbf{0.779} & 0.597 & 0.251 & 0.776 \\ \midrule
    NN2, $v \geq 2$ & 0.496 & 0.965 & 0.208 & 0.868 & NN2, 651/7 & 0.610 & \textbf{0.923} & \textbf{0.390} & \textbf{0.882} \\
    NN4, $v \geq 2$ & 0.489 & \textbf{0.966} & \textbf{0.211} & \textbf{0.871} & NN4, 651/7 & 0.616 & 0.910 & 0.356 & 0.873 \\
    NN8, $v \geq 2$ & \textbf{0.557} & 0.903 & 0.094 & 0.834 & NN8, 651/7 & \textbf{0.703} & 0.742 & 0.179 & 0.806 \\ \midrule
    NN2, $v \geq 3$ & 0.578 & 0.972 & 0.302 & 0.911 & NN2, 659 & 0.780 & \textbf{0.910} & \textbf{0.627} & \textbf{0.929} \\
    NN4, $v \geq 3$ & 0.578 & \textbf{0.973} & \textbf{0.306} & \textbf{0.914} & NN4, 659 & 0.752 & 0.894 & 0.581 & 0.910 \\
    NN8, $v \geq 3$ & \textbf{0.609} & 0.916 & 0.130 & 0.870 & NN8, 659 & \textbf{0.819} & 0.651 & 0.313 & 0.832 \\ \midrule
    NN2, $v \geq 4$ & 0.642 & \textbf{0.977} & 0.398 & 0.939 & NN2, 660/1 & 0.530 & \textbf{0.945} & \textbf{0.370} & 0.854 \\
    NN4, $v \geq 4$ & 0.647 & \textbf{0.977} & \textbf{0.401} & \textbf{0.941} & NN4, 660/1 & 0.541 & 0.940 & 0.357 & \textbf{0.862} \\
    NN8, $v \geq 4$ & \textbf{0.654} & 0.927 & 0.172 & 0.897 & NN8, 660/1 & \textbf{0.618} &  0.797 & 0.156 & 0.796 \\ \midrule
    NN2, $v \geq 5$ & 0.697 & \textbf{0.982} & \textbf{0.503} & 0.956 & \citep{tran2013}-ML & 0.37 & & 0.27 & \\ 
    NN4, $v \geq 5$ & \textbf{0.703} & 0.980 & 0.489 & \textbf{0.958} & \citep{walsh2017} & 0.96 & & 0.75 & 0.83 \\ 
    NN8, $v \geq 5$ & 0.697 & 0.936 & 0.220 & 0.919 & \citep{tran2013}-Clin. & 0.081 & & 0.129 & \\
    \bottomrule
  \end{tabular}
  \caption{Results on the test set, for which no bootstrapping has been applied. NN2/4/8 is our neural network model with, respectively, 2, 4, or 8 hidden layers; see the main text for details.  Here $v \geq n$ means that we only issue predictions for patients who have made at least $n$ visits.  CCS designations (e.g., 662) indicate that we only issue predictions for patients whose prior diagnoses includes the corresponding diagnostic codes; this is detailed in the main text.  In the lower-right corner, we quote results from the literature.  ML stands for the best machine learning model from \citep{tran2013}, while Clin. denotes the accuracy of a human clinician. For all metrics, we define the ``positive'' class to be $y=1$. Sens. = sensitivity, Spec. = specificity, Prec. = precision, AUC = area under the ROC curve.  For each metric, we have boldfaced best values within each subgroup.}
  \label{tab:allresults}
\end{table}

\section{Results and Discussion}
Table \ref{tab:allresults} shows results for the three trained networks: NN2, NN4, and NN8 (described above).  All results described in this section are test set results.  That is, we use the trained network to generate predictions for test set predictors $X^t$, after normalizing and discarding columns as described above.  Note that the prevalence of suicide attempts in the test set is $1.59\%$, close to that of the training set.

In the upper-left corner of Table \ref{tab:allresults}, we see the results for the entire test set.  NN4 has the best results in categories other than sensitivity; in this metric, the NN8 model is the clear winner.  Interestingly, the NN8 network is much more willing than NN2 or NN4 to issue a prediction of $y=1$, i.e., that a given visit is associated with a future suicide attempt.  With NN2 or NN4, poor choices of hyperparameters lead to models that predict $y=0$ always.

In the remainder of the left half of Table \ref{tab:allresults}, we have kept the trained models fixed but now evaluated their peformance on subsets of the test set.  Specifically, we set a threshold of $n$ minimum visits in a patient's record; in these $v \geq n$ results, we simply do not issue a prediction if the patient has made less than $n$ visits including the present one.  The prevalence of suicide attempts in these subsets is, respectively, $1.78\%$ ($v \geq 2$), $2.00\%$ ($v \geq 3$), $2.26\%$ ($v \geq 4$), and $2.59\%$ ($v \geq 5$).  As one would expect, the performance of all models improves as the amount of data we have about a given patient increases.  It is clear that the increases in sensitivity and precision (or positive predictive value) are not solely due to the increased prevalence of suicide attempts as $v$ increases.  The AUC values for the $n \geq 5$ models are highly encouraging.  

In the right-half of Table \ref{tab:allresults}, we have broken down the test set results by the presence of prior CCS diagnostic codes: 662 corresponds to a prior diagnosis of self-injury, 651/657 to mood and anxiety disorders, 659 to schizophrenia and psychotic disorders, and 660/661 to alcohol- and substance-related disorders.  Here we find several models with much higher sensitivity and precision (or positive predictive value) than on the test set as a whole.  This is to be expected; past history of these particular diagnostic codes is known to be associated with future suicide attempts.  Correspondingly, the prevalence of suicide attempts here is higher: $14.7\%$ (662), $7.44\%$ (651/657), $16.2\%$ (659), and $5.72\%$ (660/661).

In the lower-right corner of Table \ref{tab:allresults}, we have quoted results from the literature.  Though these studies used completely different data sets than ours, the comparisons are still interesting.  While our NN models outperform both machine and human predictions from \citet{tran2013}, they do not achieve the sensitivity or precision of \citet{walsh2017}; this latter study restricts its modeling efforts to $5167$ patients with a history of self-injury.

Our results motivate at least four areas for future work. First, we have not unpacked the features learned by the network, i.e., the mapping from the inputs $\mathbf{x}$ to the last hidden layer $\mathbf{h}^{(d)}$.  An understanding of the features will help us interpret the NN models and gauge what they have learned; it may eventually lead to the development of interpretable risk factors to predict suicide attempts.  Second, we have merely scratched the surface of neural network architectures and regularizations, leaving unexplored recent developments designed to capture the temporal nature of EHRs \citep{krishnan2017, che2017} and techniques such as dropout \citep{srivastava2014}, which may improve generalization for deeper networks.  Third, while data access/privacy issues may prevent us from running our NN models on data analyzed previously \citep{tran2013, walsh2017}, it may be possible to run others' models on our data.  Fourth, we can reframe the problem as one of quantifying how similar a given patient is to patients from the two categories.  In this way, we may be able to circumvent the class imbalance while also effectively modeling group effects.

\subsubsection*{Acknowledgments}
We acknowledge computational support from the MERCED cluster, supported by National Science Foundation grant ACI-1429783.  Research reported in this publication was supported by the National Institutes of Health under award number R15MH113108-01.  The content is solely the responsibility of the authors and does not necessarily represent the official views of the National Institutes of Health.

\newpage
\small

\bibliographystyle{plainnat}
\bibliography{bgm2017v2}

\newpage
\normalsize

\section*{Appendix}

\subsection*{Mathematical Description of the Model}
\begin{figure}[tbh]
{\centering  \includegraphics[width=1.15\linewidth]{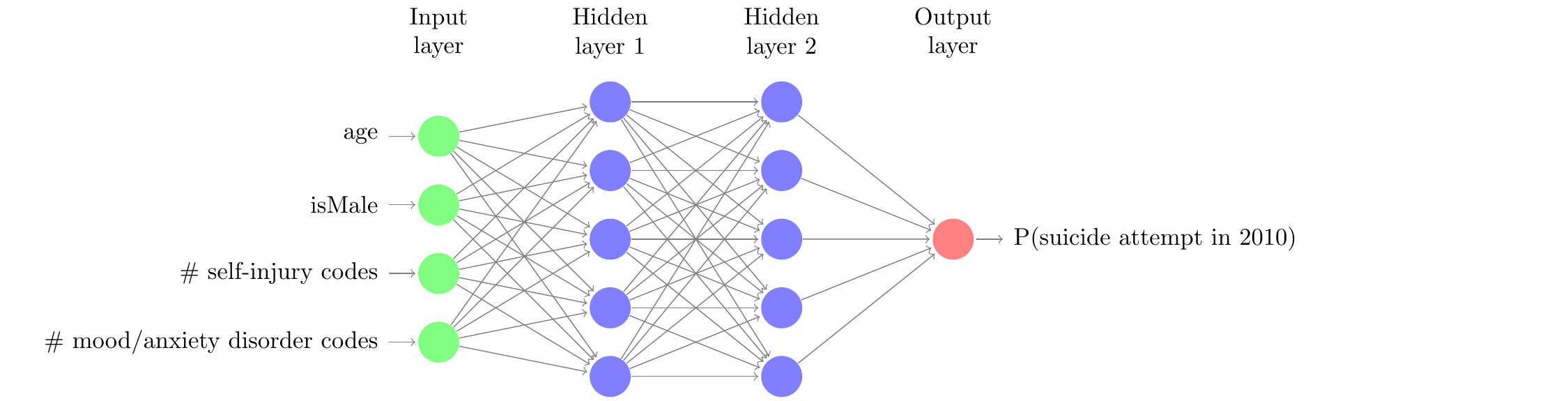} }
\caption{While our actual networks differ in the number of inputs, the number of units per layer, and the overall number of layers, the general architecture is as above.}
\label{fig:nndiag}
\end{figure}

To illustrate the architectures of our networks, we present Figure \ref{fig:nndiag}. In this diagram, information flows from left to right, from the green input nodes to the red output nodes.  Hidden layer $i$ receives a vector $\mathbf{h}^{(i-1)} \in \mathbb{R}^{n_{i-1}}$ of inputs from the layer to its left.  Suppose there are $n_i$ units or nodes in hidden layer $i$; then the trainable parameters for this layer consist of $\mathbf{W}_i$, an $n_{i-1} \times n_i$ weight matrix and $\mathbf{b}_i \in \mathbb{R}^{n_i}$, a bias vector.  With these parameters, hidden layer $i$ outputs
$$
\mathbf{h}^{(i)} = \phi(\mathbf{W}_i^T \mathbf{h}^{(i-1)} + \mathbf{b}_i)
$$
where $^T$ denotes transpose and the activation function $\phi : \mathbb{R} \to \mathbb{R}$ is applied element-wise.  

With the convention that $\mathbf{h}^{(0)} = \mathbf{x}$ where $\mathbf{x} \in \mathbb{R}^p$ is the input vector, we have completely described propagation up to the final hidden layer.  Suppose there are $d$ total hidden layers, so that the output of the final hidden layer is $\mathbf{h}^{(d)}$.  Then the output layer computes $\sigma(\mathbf{w}_o^T \mathbf{h}^{(d)} + b_o)$ with trainable weight vector $\mathbf{w}_o \in \mathbb{R}^{n_d}$ and scalar bias $b_o \in \mathbb{R}$.  Here $\sigma(z) = (1 + \exp(-z))^{-1}$ is the sigmoidal function; its output is the model's probability of a suicide attempt in 2010, i.e., $P(y=1 | x, \theta)$.

Hence our neural network computes the probability $P(y=1 | x, \theta)$ where $\theta$ stands for all trainable parameters (weights and biases).  To train the model, we find $\theta$ that maximizes the log likelihood on the training data, $\ell(\theta) = \sum_i y_i \log P(y=1 | x_i, \theta) + (1 - y_i) \log [ 1 - P(y_i=1 | x_i, \theta) ]$.  To maximize $\ell$, we employ either stochastic gradient descent (SGD)---for NN8---or SGD with momentum \citep{wilson2017}, for NN2 and NN4.  Gradients are computed using backpropagation.

In either case, we employ linear step-size decay; when we use momentum, we set the momentum parameter equal to $0.9$.  Weights for layer $i$ are randomly initialized to have mean zero and standard deviation $\sqrt{2/n_{i-1}}$; biases are initialized to zero.

All computations are carried out in Python using TensorFlow \citep{abadi2016}.  We use a rack-mounted server with two $8$-core Intel Xeon E5-2620v4 (2.1GHz) CPUs, two NVidia Tesla P100 GPUs, and $128$ GB of ECC RAM.

\newpage

\subsection*{Table of Predictors}

\begin{table}[h]
  \begin{tabular}{ll}
    \toprule
    name & type \\ \midrule
    year & numeric \\
    age & numeric \\
    zip code & numeric \\
    patient county & numeric \\
    facility ID number & numeric \\
    service year & numeric \\
    sex & categorical (4 levels) \\
    race & categorical (7 levels) \\
    insurance category & categorical (6 levels) \\
    disposition & categorical (5 levels) \\
    urban & categorical (3 levels) \\
    disposition (ED) & categorical (22 levels) \\
    facility county (ED) & categorical (55 levels) \\
    payer (ED) & categorical (20 levels) \\
    CCS diagnostic code & categorical (253 levels) \\
    number of visits (up to and including present visit) & numeric \\
    \bottomrule
  \end{tabular}
  \caption{As described in Section \ref{sect:data}, each row of our data set corresponds to a unique visit.  For each visit, our data consists of these predictors.  Here ED stands for emergency department.  Note that for each visit, we have one primary and six other CCS diagnostic codes.  For each patient, we aggregate these codes cumulatively as described in Section \ref{sect:data}.}
  \label{tab:predictors}
\end{table}
\end{document}